\documentclass[11pt]{article}
\usepackage{refcount}
\usepackage[preprint]{acl}

\usepackage{times}
\usepackage{latexsym}

\usepackage[T1]{fontenc}

\usepackage[utf8]{inputenc}

\usepackage{microtype}

\usepackage{inconsolata}

\usepackage{graphicx}
\usepackage{booktabs}

\usepackage{amsmath}
\usepackage[table]{xcolor}   
\definecolor{lightgray}{gray}{0.85}

\title{Beyond a Single Reference: Training and Evaluation with Paraphrases in Sign Language Translation\thanks{Under review.}}

\author{
Václav Javorek \\
University of West Bohemia
\And
Tomáš Železný \\
University of West Bohemia
\And
Alessa Carbo \\
Johns Hopkins University
\AND
Marek Hrúz \\
University of West Bohemia
\And
Ivan Gruber \\
University of West Bohemia
}

\begin{document}
\maketitle
\begin{abstract}
Most Sign Language Translation (SLT) corpora pair each signed utterance with a single written-language reference, despite the highly non-isomorphic relationship between sign and spoken languages, where multiple translations can be equally valid. This limitation constrains both model training and evaluation, particularly for n-gram–based metrics such as BLEU.

In this work, we investigate the use of Large Language Models to automatically generate paraphrased variants of written-language translations as synthetic alternative references for SLT. First, we compare multiple paraphrasing strategies and models using an adapted ParaScore metric. Second, we study the impact of paraphrases on both training and evaluation of the pose-based T5 model on the YouTubeASL and How2Sign datasets.

Our results show that naively incorporating paraphrases during training does not improve translation performance and can even be detrimental. In contrast, using paraphrases during evaluation leads to higher automatic scores and better alignment with human judgments. To formalize this observation, we introduce  $\text{BLEU}_{\text{para}}$, an extension of BLEU that evaluates translations against multiple paraphrased references. Human evaluation confirms that  $\text{BLEU}_{\text{para}}$ correlates stronger with perceived translation quality. We release all generated paraphrases, generation and evaluation code to support reproducible and more reliable evaluation of SLT systems.
\end{abstract}

\section{Introduction}
Automatic Sign Language Translation (SLT) has recently made rapid progress due to advances in Vision Language Models and the availability of large-scale video–text corpora. Despite these developments, a persistent limitation in current SLT datasets is the scarcity of translation variability: each sign language utterance is typically paired with a single weakly aligned written-language reference translation. This stands in contrast to spoken language machine translation, where corpora frequently provide multiple reference translations or explicitly annotate translation alternatives, capturing the natural diversity of phrasing, register, and information structure that exists in human language. Such variation is crucial for both training and evaluation, where metrics such as BLEU are known to be sensitive to the availability of reference translations due to their n-gram matching nature.

Sign languages pose an even stronger requirement for translation diversity. The mapping between visual expression and written language is highly non-isomorphic: a single sign language sentence can be translated into several equally valid spoken language realizations, differing in word order, degree of explicitness, and syntactic structure. Relying on a single reference translation therefore risks over-constraining model training and underestimating translation quality during evaluation.

In this work, we explore whether Large Language Models (LLMs) can be used to systematically enrich SLT corpora with automatically generated paraphrases of written language translations. Our goal is twofold:
(1) to study whether synthetic generation of translation variants (i.e. paraphrasing) can improve the robustness of sequence-to-sequence SLT models such as T5, and
(2) to examine how these paraphrases affect evaluation, comparing BLEU scores against human evaluation.

By viewing paraphrases as alternative translations, analogous to multi-reference machine translation setups, we aim to bridge principles from classical natural language processing (NLP) with the specific challenges of SLT. 

\section{Related work}

\paragraph{Sign Language Translation} Recent progress in SLTn have been driven by increasingly powerful multimodal and transformer-based architectures. A commonly used baseline architecture consists of a keypoint extractor (e.g. MMPose~\cite{sengupta2020mm} or MediaPipe~\cite{lugaresi2019mediapipe}), followed by transformer~\cite{vaswani2017attention}. This approach became an effective way to perform experiments and provide benchmark results for new datasets~\cite{yasl, 9667087, tanzer2025youtubesl, tanzer-2025-fleurs, alhejab2025saudi}. Recent state-of-the-art methods extend this approach by adding additional modules. GloFE~\cite{lin2023gloss} employs visual backbone based on graph convolutional networks (GCN) to more effectively process pose features, followed by a transformer encoder–decoder architecture that learns robust gloss-level representations through contrastive learning. Uni-Sign~\cite{scale88uni-unisign} similarly uses GCN for pose feature encoding and further incorporates a fusion module that integrates RGB information in cases where keypoint extraction is failing. 
SignBERT+~\cite{hu2023signbert+} extends BERT-style pretraining~\cite{devlin2019bert} to sign language data, learning contextualized representations that capture temporal dependencies within signing sequences and can be transferred to downstream tasks such as isolated and continuous sign language recognition.
LITFIC~\cite{jang2025lost-litfic} leverages a LLaMA-based~\cite{touvron2023llama} large language model that incorporates contextual and meta-information, including pseudo-gloss cues, to improve the translation quality. By integrating these diverse sources of information, the model can better capture discourse-level context and disambiguate signs.

\paragraph{Paraphrasing} Paraphrasing - generating semantically equivalent expressions with varied surface forms - has been extensively studied for both its theoretical properties and practical applications in NLP. Bhagat \& Hovy~\cite{bhagat-hovy} established that quality paraphrases balance semantic preservation against lexical and syntactic diversity, a tension later formalized in evaluation metrics like ParaScore~\cite{shen2022parascore}, which combines BERTScore-based~\cite{zhang2020bertscore} semantic similarity with explicit diversity modeling. Neural paraphrase generation has evolved from seq2seq models to transformer-based approaches fine-tuned on large-scale datasets like ParaNMT-50M~\cite{adversarial}. In machine translation, paraphrasing serves as a key data augmentation strategy, particularly for low-resource settings where back-translation~\cite{sennrich-etal-2016-neural} and lexical substitution have yielded substantial BLEU improvements. For SLT specifically, paraphrasing is particularly relevant where due to he structural divergence between signed and written languages means a single signed utterance often admits multiple valid translations, yet datasets typically provide only one translation.

\paragraph{Data} In this work, we focus on American Sign Language, namely by using two widespread datasets - YoutubeASL~\cite{yasl} and How2Sign~\cite{how2sign}. YouTubeASL is a large-scale dataset of American Sign Language (ASL) collected from publicly available YouTube videos. It contains over 610,000 video samples (equivalent to about 1,000 hours) and features 2,500 unique signers from diverse backgrounds. Videos cover various domains, such as education, personal vlogs, and conversational ASL, making the dataset representative of a wide range of signing contexts.

How2Sign is a smaller but well-annotated dataset, originally developed to study sign translation under multiple angles/views. It contains 35,000 signing samples, amounting to approximately 80 hours of ASL content, produced by roughly 11 signers in a controlled environment with stable lighting and fixed backgrounds. Although the dataset’s primary focus is translating from spoken English to ASL, it is widely adopted for other tasks in the sign language processing community, including sign language translation. One challenge with using How2Sign lies in the presence of translationese effects, a phenomenon where translated text can sound artificial or lack natural linguistic flow~\cite{graham-etal-2020-statistical}. Despite this limitation, How2Sign serves as an important benchmark for evaluating model performance. Its controlled nature provides consistency that complements the more variable, in-the-wild data from YouTubeASL.

\section{Methods}

\subsection{Sign Language Translation}\label{sec:slt}

For our experiments, we adopt an existing SLT framework that uses a modified T5 encoder–decoder transformer to process pose-based inputs and includes comprehensive data preprocessing, such as keypoint extraction, normalization, interpolation, and augmentation~\cite{zelezny2025exploring}. We adopt this existing setup because it provides already ready‑to‑use codebase and preprocessed YouTubeASL dataset~\cite{11234/1-5898}, including extracted keypoints and corresponding annotations. This allows us to focus on evaluating our novel contributions without re‑implementing core components such as data handling and model configuration. The publicly available repository accompanying that work includes all preprocessing pipelines, model training routines, and evaluation scripts, ensuring reproducibility and consistent comparison across experiments.

\subsection{Paraphrasing and ParaScore}

Paraphrasing plays a vital role in modern natural language processing and machine translation, making the quality assessment of paraphrases essential for these applications. In the context of SLT, where the mapping between visual and written modalities admits multiple valid realizations, paraphrases can serve as synthetic valid alternative references. However, not all automatically generated paraphrases will be equally useful: a paraphrase that diverges too far semantically fails to represent the source meaning, while one that is too lexically similar to the original provides little additional signal. A successful paraphrase must balance two critical criteria: semantic similarity and lexical divergence. To evaluate this balance, we employ an adapted version of the ParaScore metric~\cite{shen2022parascore}, which effectively combines semantic preservation with linguistic diversity through the use of both language models and traditional linguistic measures.

Unlike most existing metrics that struggle to adequately balance semantic similarity and lexical divergence, this approach implements a max function for the semantic similarity component between an input sequence and its candidate rephrasement. The semantic similarity component utilizes BERTScore~\cite{zhang2020bertscore}, which employs the following process: it first computes contextual embeddings for each token in both sentences, then calculates cosine similarities between tokens in the two sentences, and finally employs a matching strategy to compute precision, recall, and F1 score.

For measuring lexical divergence, we implement the Levenshtein distance~\cite{levenshtein1965binary}, which calculates the minimum number of single-character edits (insertions, deletions, or substitutions) required to transform one word into another. This distance is normalized by dividing by the length of the longer string, resulting in a value between 0 and 1, where 0 indicates identical strings and 1 represents completely different strings.

The metric incorporates two hyperparameters: $\gamma$ and $\omega$. $\gamma$, set to a default value of 0.35, serves two purposes: it caps the maximum divergence score and introduces non-linearity to the scaling of the divergence score. This means that edit distances above 35\% of the longer string's length are treated equivalently. $\omega$, with a default value of 0.5, is used in the final ParaScore calculation to weight the divergence. Formally, given an input $x$ and a candidate paraphrase $\hat{x}$, the metric is defined as:

\begin{equation}
\resizebox{\linewidth}{!}{$
\text{ParaScore}(x, \hat{x}) =
\frac{\text{BERTScore}(x, \hat{x}) + \omega \cdot \min(\text{NLD}(x, \hat{x}), \gamma)}
{1 + (\omega \cdot \gamma)}
$}
\end{equation}

\noindent where $\text{NLD}$ is the Normalized Levenshtein Distance. With these parameters, the similarity score can contribute up to 1 to the final score, while the divergence score can contribute up to 0.175 (i.e., $0.5 \times 0.35$). Although this creates a maximum possible score of 1.175, we normalize the final score by dividing by this maximum (represented by the denominator) to achieve a standardized scale.

\section{Comparison of Paraphrasing Methods}

\begin{figure*}[ht]
  \centering
  \includegraphics[width=\linewidth]{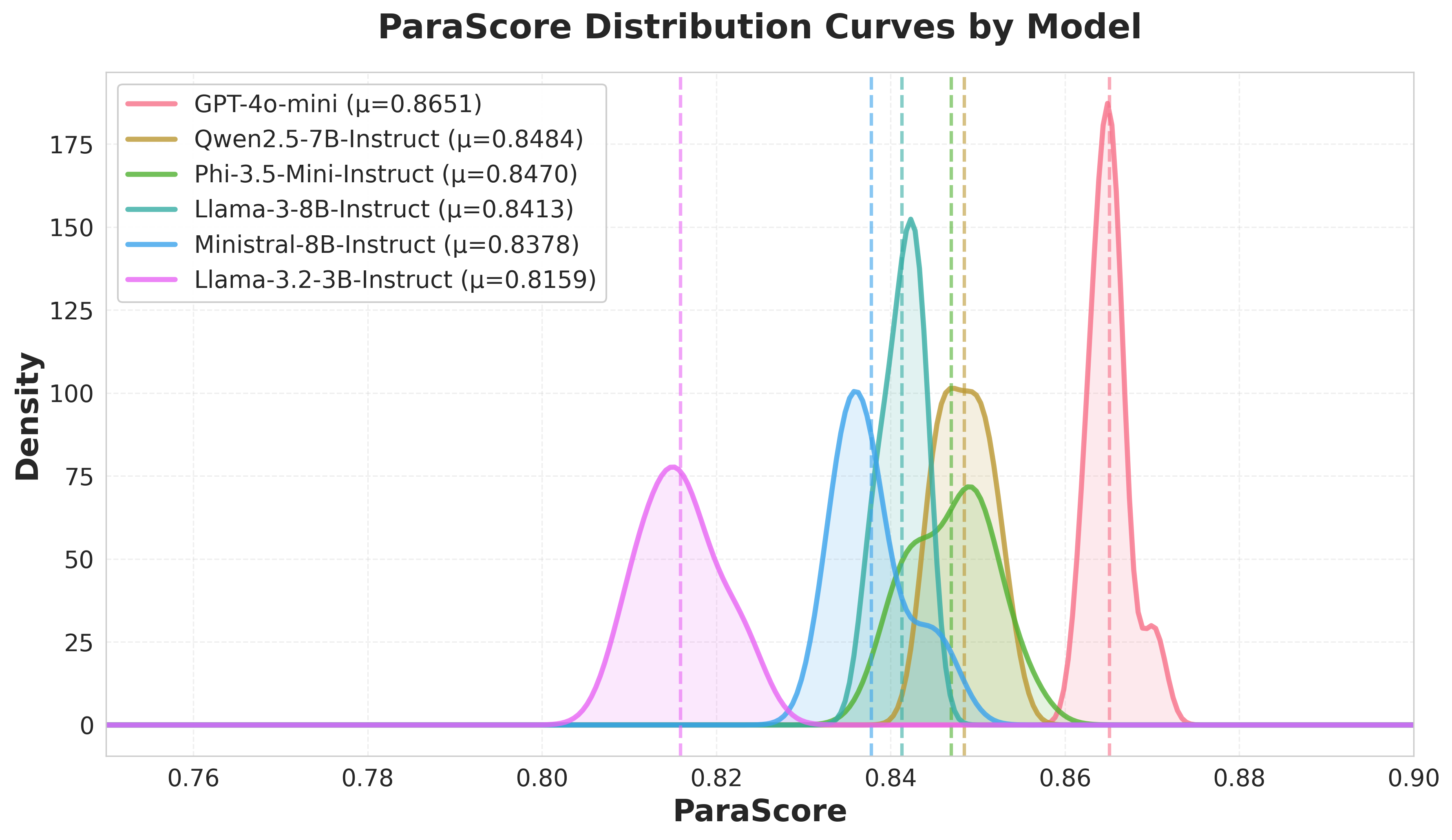}
  \caption{\textbf{ParaScore distributions for paraphrases generated by different LLMs.} Kernel density estimates over all generated paraphrases (same prompt and decoding for all models). Dashed vertical lines mark per-model means (\(\mu\); shown in the legend). Higher scores indicate paraphrases that better preserve meaning while avoiding near-copies, enabling a direct quality comparison across paraphrasing models.}
  \label{LLM_comparison}
\end{figure*}

\begin{figure*}[ht]
  \centering
  \includegraphics[width=0.9\linewidth]{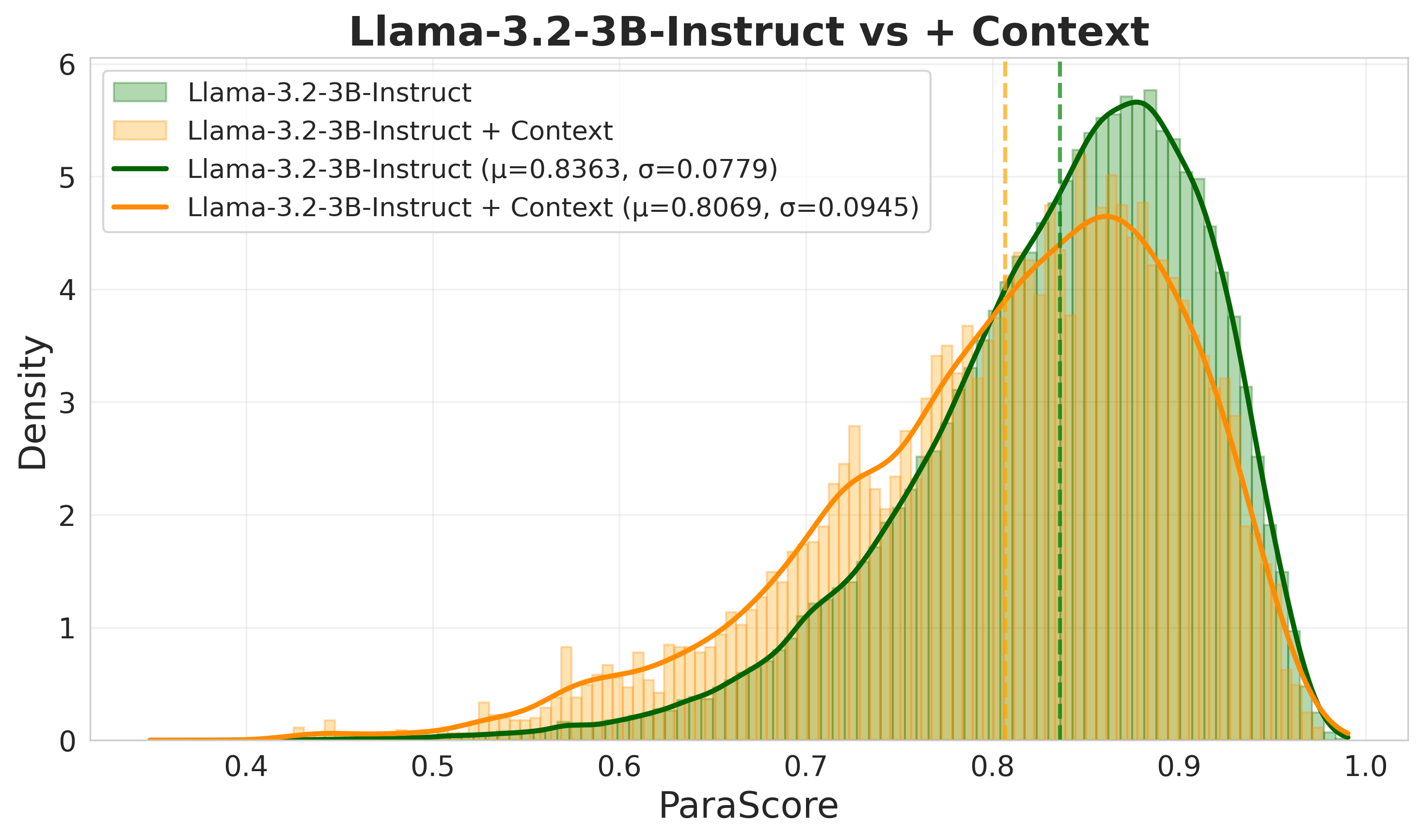}
  \caption{\textbf{Effect of adding video-level context to the paraphrasing prompt.} ParaScore distributions for Llama-3.2-3B-Instruct with sentence-only prompting versus prompting augmented with short preceding context from the same video clip. Dashed lines denote means (\(\mu\)); providing context shifts the distribution left and increases variance, suggesting that extra discourse information can encourage looser rewrites that more often drift from the original meaning under our ParaScore criterion.}
  \label{context}
\end{figure*}

\begin{figure*}[ht]
  \centering
  \includegraphics[width=0.9\linewidth]{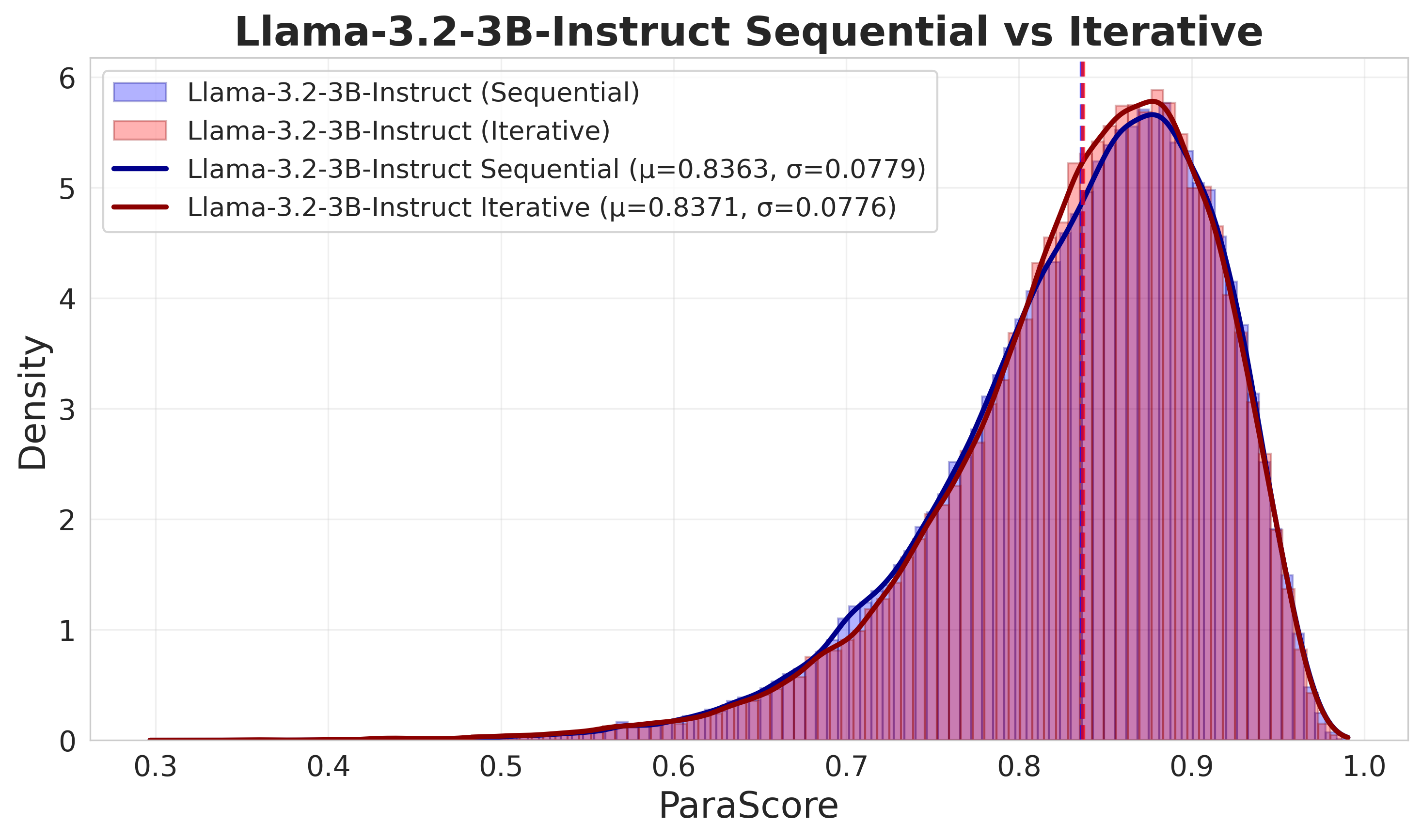}
  \caption{\textbf{Sequential vs.\ iterative paraphrase generation.} ParaScore distributions for Llama-3.2-3B-Instruct when producing five paraphrases in a single call (\textit{sequential}) versus generating them one-by-one with feedback from earlier outputs (\textit{iterative}). Histograms and density curves largely overlap, and the mean/variance (dashed lines; \(\mu,\sigma\) in legend) are nearly identical, indicating that iterative prompting does not materially change paraphrase quality in our setting.}
  \label{seq_iter}
\end{figure*}

In this section we compare paraphrasing models under a single, controlled generation pipeline. For each reference English sentence \(x\) in our SLT corpora, we ask a candidate LLM to produce \(K{=}5\) meaning-preserving rewrites \(\{\hat{x}_k\}_{k=1}^K\) using the same system instruction across models: \texttt{You are a helpful assistant that rephrases a given sentence in K ways, each on its own line. Try to be semantically consistent and output nothing else than these sentences. Sentence: <x> Paraphrases:}. To keep outputs comparable, we use consistent decoding (sampling; temperature \(=0.7\), top-\(p{=}0.95\)). We then normalize generations by stripping list markers/bullets and occasional boilerplate text (e.g., ``Here are \ldots''); sentences outside 4--30 words are skipped, and any generation that does not yield exactly \(K\) lines is treated as missing. The resulting per-model paraphrase sets are scored with ParaScore and used for the analyses in FiguresFigures~\ref{LLM_comparison}, \ref{context}, \ref{seq_iter} and the subsequent experiments.

Through careful manual evaluation of paraphrases at various percentile levels (0.25, 0.50, 0.75), we established a quality threshold of 0.7. This threshold serves as a reliable indicator for distinguishing high-quality paraphrases from lower-quality ones.

For paraphrasing ASL translations, we use a wide range modern Large Language Models. Their results are compared in Figure~\ref{LLM_comparison}.

Because SLT is currently done mostly at sentence-level, the given sentence is usually a few seconds clip, extracted from a longer video, lacking context, which was proven particularly problematic as Sign Languages are very context-dependent~\cite{jang2025lost-litfic}. 

As a separate experiment, we augment the paraphrasing prompt with lightweight \emph{video-level textual context}. For each target sentence, we add the preceding reference sentence(s) from the same video clip, and reset the context when moving to a new video. This context is provided as part of the instruction while keeping the target sentence unchanged, allowing us to isolate whether discourse information (co-reference, ellipsis, topic continuity) improves meaning-preserving paraphrases compared to sentence-only prompting. However, in our experimental setting the ParaScore is worse for the paraphrases generated with context (see Figure~\ref{context}). This is something we are planning to explore further in our future work.

Another separate experimental setting tackles the use of prompting to gain multiple paraphrases of the same reference translation sentence. We can either use \textbf{sequential} (ask for n paraphrases at once in a single prompt) or \textbf{iterative} prompting (ask for a paraphrase n times). Sequential prompting proved to generate paraphrases with a little higher average ParaScore (see Figure~\ref{seq_iter}) and for a lower computation cost.

In our comparison, we applied our adapted ParaScore metric to each dataset, analyzing both individual paraphrases and the average scores for each caption. The analysis revealed that GPT-4o-mini paraphrasing achieved the highest average ParaScore; therefore, its generated paraphrase sets for YouTubeASL and How2Sign are used in the subsequent training and evaluation experiments. Notably, we observed that adding context to both sequential and iterative prompting in LLaMA led to significant performance degradation.

\section{Paraphrases for Training and Evaluation}\label{sec:slttraining}
In this section, we investigate the impact of paraphrases on SLT with two main objectives: (1) To examine whether incorporating paraphrased target sentences during training improves SLT performance. (2) To assess whether using paraphrases can lead to improved automatic evaluation quality. 

Different paraphrase-based training and evaluation strategies are compared in a controlled experimental setting. Evaluation using paraphrased references is particularly interesting, as higher alignment with human judgment would indicate their potential usefulness; the analysis of human judgments is discussed in Section~\ref{sec:humaneval}.

\subsection{Training Setup}
We follow the default training configuration of the adopted framework (see~\ref{sec:slt}), using the same hyperparameter values and preprocessing settings. Experiments are conducted on the pose-based YouTubeASL dataset with pre-extracted keypoints, adopting the provided 90:10 train–val split. For the How2Sign dataset, we use the original dataset split. Input pose sequences are normalized using SignSpace~\cite{zelezny2025exploring} normalization, with missing keypoints filled using a fixed value of $-10$. Models are pretrained on YouTubeASL dataset with an effective batch size of 256 and a learning rate of 0.0004, employing a warm-up phase of 5,000 steps. The models are trained until convergence, i. e. about 400,000 iteration steps. We then finetune the models on How2Sign dataset using the same setting until convergence, i. e. about 10,000 steps. No data augmentation techniques are applied during nor of the training stages.

For evaluation, we report BLEU scores computed using the \texttt{sacrebleu} library v2.4.3\footnote{\url{https://pypi.org/project/sacrebleu/2.4.3/}}, ROUGE-L scores from \texttt{evaluate} library\footnote{\url{https://github.com/huggingface/evaluate}}, and BLEURT scores using the \texttt{BLEURT-20} checkpoint from the official repository\footnote{\url{https://github.com/google-research/bleurt}}.

\subsection{Experiments}

First, the model is pretrained on the YouTube ASL dataset without paraphrases. It is then fine-tuned on the How2Sign dataset using different types of paraphrase-based training strategies. The resulting model checkpoints are then evaluated using the How2Sign test set. To ensure consistency of results, each experiment is run three times with different random seeds, and all reported results correspond to the average performance across these runs. Complete results are provided in the supplementary material.

We consider three training configurations. First, we train a baseline model using only the canonical target sentence without paraphrases. Second, we train a model where the target sentence is randomly sampled from up to six alternatives (one canonical reference and five paraphrases) for each training instance. Third, we compute the training loss for all available paraphrases and backpropagate the gradients from the paraphrase yielding the minimum loss for a given instance.

In the paraphrase-based evaluation setting, we calculate metrics for all available paraphrases of each test instance and select the highest-scoring paraphrase as the final result. As shown in Table~\ref{tab:train1}, models trained without paraphrases outperform those trained with paraphrased targets. This suggests that exposure to multiple paraphrased targets during training may introduce ambiguity that negatively impacts the performance. As expected, evaluation scores obtained using paraphrased references are generally higher. However, the relationship between these automatic metrics and human judgment is analyzed in the following section.

\begin{table}[]
\centering
\begin{tabular}{l|ccc}
\toprule
Par. mode & BLEU-4 & BLEURT & Rouge-L\\ \midrule
\multicolumn{4}{c}{\cellcolor{lightgray}{Evaluation without paraphrases}} \\ \midrule
No par. & \textbf{7.46} & \textbf{0.43} & \textbf{0.29} \\
Random & 6.17 & \textbf{0.43} & 0.26 \\
Min loss & 6.49 & \textbf{0.43} & 0.27 \\ \midrule
\multicolumn{4}{c}{\cellcolor{lightgray}{Evaluation with paraphrases}} \\ \midrule
No par. & \textbf{8.86} & 0.44 & \textbf{0.31} \\
Random & 7.95 & \textbf{0.45} & 0.30 \\
Min loss & 8.00 & 0.44 & 0.30 \\
\bottomrule
\end{tabular}
\caption{Performance comparison of different paraphrase-based training strategies on the How2Sign test set using BLEU-4, BLEURT, and ROUGE-L metrics. The checkpoints were evaluated both without and with paraphrases, where metrics are computed for all available paraphrases of each test instance and the highest-scoring paraphrase is selected. \texttt{No~par.} denotes training with only canonical targets, \texttt{Random} indicates random sampling from available paraphrases during training, and \texttt{Min~loss} selects the paraphrase yielding the minimum training loss per instance. For the \texttt{Min~loss} configuration, a paraphrased target was selected in approximately 60\% of training instances.}
\label{tab:train1}
\end{table}

\section{Human Evaluation}\label{sec:humaneval}

To address prior findings~\cite{jang2025lost-litfic}, that show concerning limitations of the standardized metrics in SLT, usually focused on WER and translation length, we propose the $\text{BLEU}_{\text{para}}$ metric, enhancing the standard BLEU4~\cite{Papineni2002BleuAM} metric with 5 paraphrases of each reference sentence in the dataset, fixing any word order or synonym issues. To enable its use, we publish the LLM-generated paraphrased versions of How2Sign~\cite{how2sign} and YoutubeASL~\cite{yasl} alongside this paper
.

To prove that this metric improves SLT evaluation over its non-paraphrased counterpart, we have conducted a Human Evaluation experiment, using the exact same protocol as~\cite{jang2025lost-litfic}. In addition to reporting correlations over all items, we explicitly probe \emph{extreme} cases where reference-matching metrics are typically the most brittle: we select the 48 examples whose canonical-reference BLEU is either very low ($<5$) or very high ($>15$). This subset captures both clear failures and near-verbatim matches, and highlights whether $\text{BLEU}_{\text{para}}$ better tracks human judgments in outlier translations.

The results in Table \ref{human_eval} show that human perception of model capabilities correlates better with $\text{BLEU}_{\text{para}}$, especially in outlier cases (48 translations with $\text{BLEU}{<}5$ or $>{}15$).

\begin{table}[t]
\centering
\small
\setlength{\tabcolsep}{4pt}
\renewcommand{\arraystretch}{1.05}
\begin{tabular}{lccc}
\toprule
\textbf{Metric} & \textbf{Pearson $r$} & \textbf{Spearman $\rho$} & \textbf{$\rho$ (extremes)} \\
\midrule
$\text{BLEU}$  & 0.688 & 0.657 & 0.578 \\
$\text{BLEU}_{\text{para}}$ & \textbf{0.697} & \textbf{0.685} & \textbf{0.659} \\
\bottomrule
\end{tabular}
\caption{Correlation between mean human ratings (0--5; 70 sentences, 6 annotators; mean $=1.45\pm1.14$) and automatic metrics. ``Extremes'' uses translations with $\text{BLEU}{<}5$ or $>{}15$.}
\label{human_eval}
\end{table}

\section{Discussion}
Our comparison of paraphrasing methods reveals substantial variability among individual LLMs in their ability to generate high-quality paraphrases for synthetic translation data. Among the evaluated models, GPT-4o outperforms the other methods by a considerable margin. We assume the trend to be improving, with the expectation that commercial models will be slightly more capable than publicly available models. It is therefore likely that the results achieved by GPT-4o in our experiments may already be surpassed by newer versions of commercial models, such as GPT-5 or Gemini 3. Nevertheless, because these experiments are financially demanding and the primary goal of this paper is not to achieve state-of-the-art results, but rather to explore a general concept, we leave additional experiments for future research.

Incorporating paraphrases directly as data augmentation during model training does not yield measurable performance gains, suggesting that naive output-level augmentation is ineffective. While introducing multiple paraphrase variants into the loss computation could potentially improve learning, this approach significantly increases training time in the current setup. More selective strategies, such as hard example mining, may therefore be a more promising direction for future research.

Prior work has demonstrated that current standardized metrics, such as BLEU and ROUGE, rely heavily on word-level matching strategies, which correlate with human perception only weakly. To address this limitation, we propose a novel metric $\text{BLEU}_{\text{para}}$. Human evaluation confirms that the correlation is improved by employing this new metric instead of the standard BLEU metric. Additionally, $\text{BLEU}_{\text{para}}$ partially alleviates issues related to synonyms and word order. Its main disadvantage is the necessity to have paraphrases for each test sample. To reduce this burden and support reproducibility, we publish all paraphrases, including the codes for their creation, in our GitHub repository
. 

\section{Conclusion}
This paper experiments with the use of paraphrases for training and evaluation. We first compare the capabilities of different large language models in the task of paraphrase creation. Then we utilize these paraphrases for both training and evaluation. While the training with paraphrases, using basic ideas for its incorporation, does not bring any measurable improvements, we would like to explore more complex protocols in our future research. 

On the other hand, the evaluation using paraphrases provides results better correlated with human perception of model capabilities. To reflect the usage of paraphrases during the evaluation step, we propose a novel metric $\text{BLEU}_{\text{para}}$. We argue that this metric is more suitable for the evaluation of methods for SLT than the standard metrics like BLEU or ROUGE. All the codes and paraphrases are publicly available in our GitHub repository
. 

In our future work, we would like to explore different protocols for paraphrase creation more in depth, especially the utilization of context. Additionally, we would like to test if the proposed metric $\text{BLEU}_{\text{para}}$ generalizes well also for other sign languages other than American Sign Language. 


\section*{Limitations}

The main limitation of this research is scope and reproducibility. Our experiments are confined to a small set of datasets and one translation setting, so the findings may not transfer to other sign languages, domains, or annotation styles. Results are also sensitive to the choice of paraphrasing model, decoding settings, and prompts; in particular, using proprietary LLMs and stochastic sampling can make exact reproduction difficult and may change over time as providers update models. Finally, compute and cost constraints limited the number of paraphrases per reference, the breadth of hyperparameter sweeps, and the scale of human evaluation, so some effects (especially small gains between strong models) may be underpowered and should be revalidated on larger samples.

Our work is limited by the technical properties of sign language video data and its alignment with translations. Variations in resolution, frame rate, motion blur, camera framing, lighting, occlusions, and background contrast can reduce the visibility of important manual and non-manual cues. Video segments are often temporally incomplete or misaligned with their translations, lacking natural onset and offset phases, which introduces noise and complicates training and evaluation. These factors affect both the reliability of model learning and the interpretability of automatic metrics, including experiments with paraphrased targets and evaluation references.


\bibliography{custom}

\appendix

\section{Supplementary material}
\label{sec:supplementary}
\subsection{SLT Evaluation Details}
In order to preserve consistency of our experiments, we run each of our experiments with three different seeds, as discussed in Section~\ref{sec:slttraining}. Detailed results for all of our runs and all metrics we evaluated on are reported in Table~\ref{tab:detailed_results}.


\begin{table*}[]
\centering
\begin{tabular}{l|r|cccccc}
\toprule
Paraphrase mode & Seed & BLEU-1  & BLEU-2  & BLEU-3  & BLEU-4 & BLEURT & Rouge-L\\ \midrule
\multicolumn{8}{c}{\cellcolor{lightgray}{Evaluation without paraphrases}} \\ \midrule
No paraphrases & 42 & 28.34 & 16.38 & 10.76 & 7.44 & \textbf{0.43} & \textbf{0.29} \\
No paraphrases &  0 & 27.16 & 15.90 & 10.59 & 7.45 & \textbf{0.43} & \textbf{0.29} \\
No paraphrases &  1 & \textbf{28.69} & \textbf{16.50} & \textbf{10.83} & \textbf{7.48} & 0.42 & \textbf{0.29} \\\hline
Random & 42 & 24.80 & 13.97 & 9.02  & 6.15 & \textbf{0.43} & 0.27 \\
Random &  0 & 26.64 & 14.78 & 9.37  & 6.34 & 0.42 & 0.26 \\
Random &  1 & 24.48 & 13.81 & 8.86  & 6.02 & \textbf{0.43} & 0.26 \\\hline
Min loss    & 42 & 26.79 & 15.14 & 9.86  & 6.82 & 0.42 & 0.27 \\
Min loss    &  0 & 22.23 & 12.78 & 8.35  & 5.76 & \textbf{0.43} & 0.28 \\
Min loss    &  1 & 26.87 & 15.41 & 10.03 & 6.89 & \textbf{0.43} & 0.27 \\\midrule
\multicolumn{8}{c}{\cellcolor{lightgray}{Evaluation with paraphrases}} \\ \midrule
No paraphrases & 42 & 33.75 & 19.78 & 12.92 & 8.85 & 0.44 & \textbf{0.31} \\
No paraphrases &  0 & 32.43 & 19.20 & 12.71 & 8.85 & \textbf{0.45} & \textbf{0.31} \\
No paraphrases &  1 & \textbf{34.06} & \textbf{19.91} & \textbf{12.99} & \textbf{8.88} & 0.44 & \textbf{0.31} \\\hline
Random & 42 & 30.33 & 17.75 & 11.75 & 7.85 & \textbf{0.45} & 0.30 \\
Random &  0 & 33.03 & 19.15 & 12.30 & 8.30 & 0.44 & 0.30 \\
Random &  1 & 29.93 & 17.49 & 11.34 & 7.71 & \textbf{0.45} & 0.29 \\\hline
Min loss    & 42 & 32.51 & 18.86 & 12.29 & 8.43 & 0.44 & 0.30 \\
Min loss    &  0 & 26.76 & 15.72 & 10.90 & 7.06 & \textbf{0.45} & 0.30 \\
Min loss    &  1 & 32.42 & 19.09 & 12.44 & 8.51 & 0.44 & 0.30 \\
\bottomrule
\end{tabular}
\caption{Performance of our models trained with different seeds and different paraphrase-based training strategies on the How2Sign test set. The checkpoints were evaluated both without and with paraphrases, where metrics are computed for all available paraphrases of each test instance and the highest-scoring paraphrase is selected. \texttt{No~par.} denotes training with only canonical targets, \texttt{Random} indicates random sampling from available paraphrases during training, and \texttt{Min~loss} selects the paraphrase yielding the minimum training loss per instance. For the \texttt{Min~loss} configuration, a paraphrased target was selected in approximately 60\% of training instances.}
\label{tab:detailed_results}
\end{table*}




\end{document}